\documentclass[10pt,twocolumn,letterpaper]{article}

\usepackage{cvpr}
\usepackage{times}
\usepackage{epsfig}
\usepackage{graphicx}
\usepackage{amsmath}
\usepackage{amssymb}

\usepackage[latin9]{inputenc}
\usepackage{amsmath,amssymb,amsfonts}
\usepackage{times}
\usepackage{graphicx}
\usepackage{epsfig}
\usepackage{multirow}
\usepackage{booktabs}
\usepackage{algorithmic}
\usepackage{siunitx}
\usepackage[norule]{footmisc}
\usepackage{cite}
\usepackage{algorithmic}
\usepackage{textcomp}
\usepackage{xcolor}
\usepackage{float} 
\usepackage{hyperref}

\graphicspath{{Images/}}
\DeclareGraphicsExtensions{.eps,.png,.pdf,.jpg,.jpeg,.JPG}

\def\BibTeX{{\rm B\kern-.05em{\sc i\kern-.025em b}\kern-.08em
    T\kern-.1667em\lower.7ex\hbox{E}\kern-.125emX}}

\begin{document}


\title{A Dataless FaceSwap Detection Approach Using Synthetic Images. \\

}

\author{Anubhav Jain, Nasir Memon, Julian Togelius\\
New York University\\
{\tt\small \{aj3281, nm1214, jt125\}@nyu.edu}
}

\maketitle
\thispagestyle{empty}

\begin{abstract}

Face swapping technology used to create "Deepfakes" has advanced significantly over the past few years and now enables us to create realistic facial manipulations. Current deep learning algorithms to detect deepfakes have shown promising results, however, they require large amounts of training data, and as we show they are biased towards a particular ethnicity. We propose a deepfake detection methodology that eliminates the need for any real data by making use of synthetically generated data using StyleGAN3. This not only performs at par with the traditional training methodology of using real data but it shows better generalization capabilities when finetuned with a small amount of real data. Furthermore, this also reduces biases created by facial image datasets that might have sparse data from particular ethnicities. To promote reproducibility the code base has been made publicly available \footnote{\href{https://github.com/anubhav1997/youneednodataset}{https://github.com/anubhav1997/youneednodataset}} .

\end{abstract}

\section{Introduction}

Numerous approaches have been proposed to create photo-realistic deepfakes in recent years. With every new approach that is proposed, the quality of the deepfakes is getting better; we will likely eventually reach a point where humans are unlikely to distinguish them from real images or video\cite{rossler2019faceforensics++}. Most current approaches make use of face-swapping models where the face of a person is replaced with that of another in an image or video. While current approaches to detecting deepfakes have been promising they require a large dataset of facial images.  This can be challenging as publicly available datasets have extremely strict licenses or in some cases do not have the rights for use of the data cleared - such as the FakeAVCeleb, FF++, Celeb-DF, and UADFV to name a few. Some of these datasets have directly used videos from YouTube or other such public spaces without obtaining proper consent from the people that appear in the videos. Published datasets that have claimed to use images of celebrities scraped from public forums have often included data from "ordinary" individuals \cite{ftwhoseusingyourface} and have hence been discontinued due to such reasons \cite{microsoftdetetesdataset}. This leads to a need to move away from such approaches that require large amounts of privacy-sensitive data.

\begin{figure}[t]
\centering
\subfloat{\includegraphics[width=0.29\columnwidth]{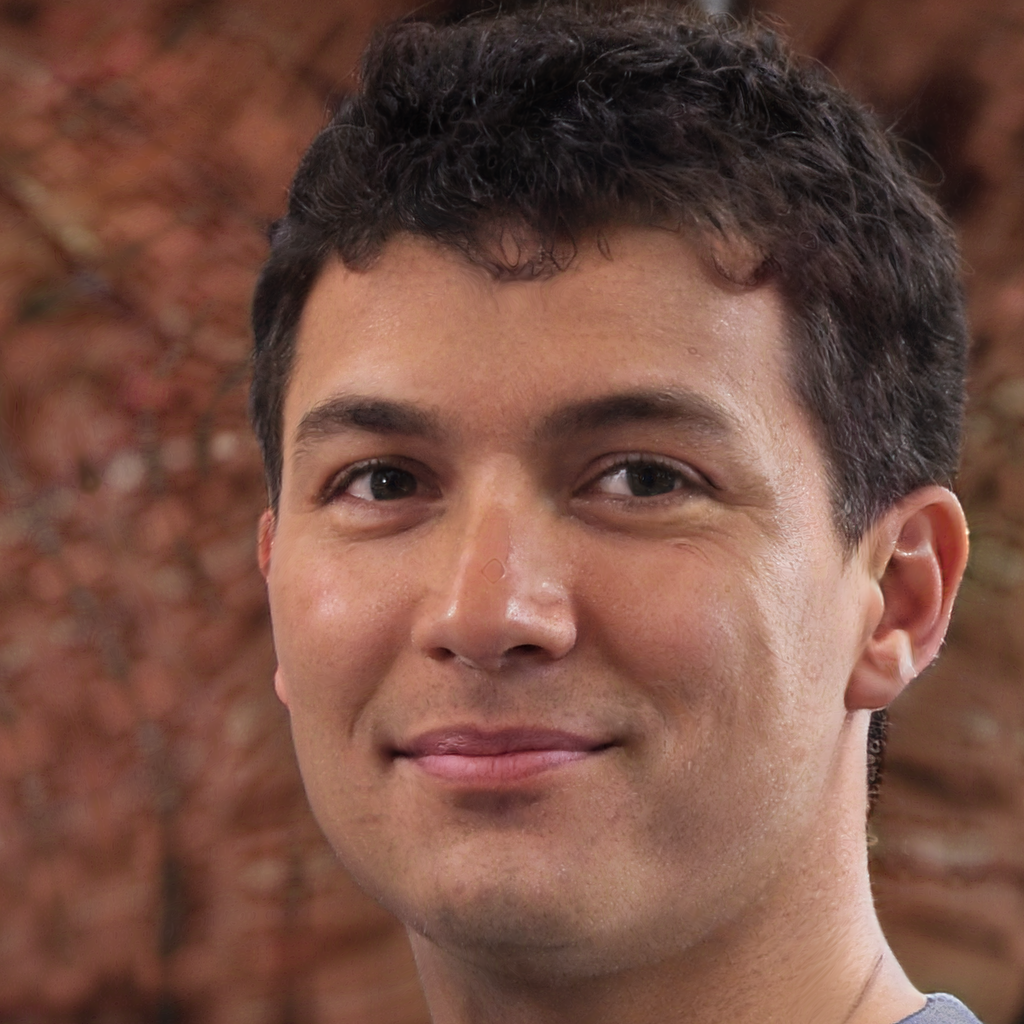}}
~
\subfloat{\includegraphics[width=0.29\columnwidth]{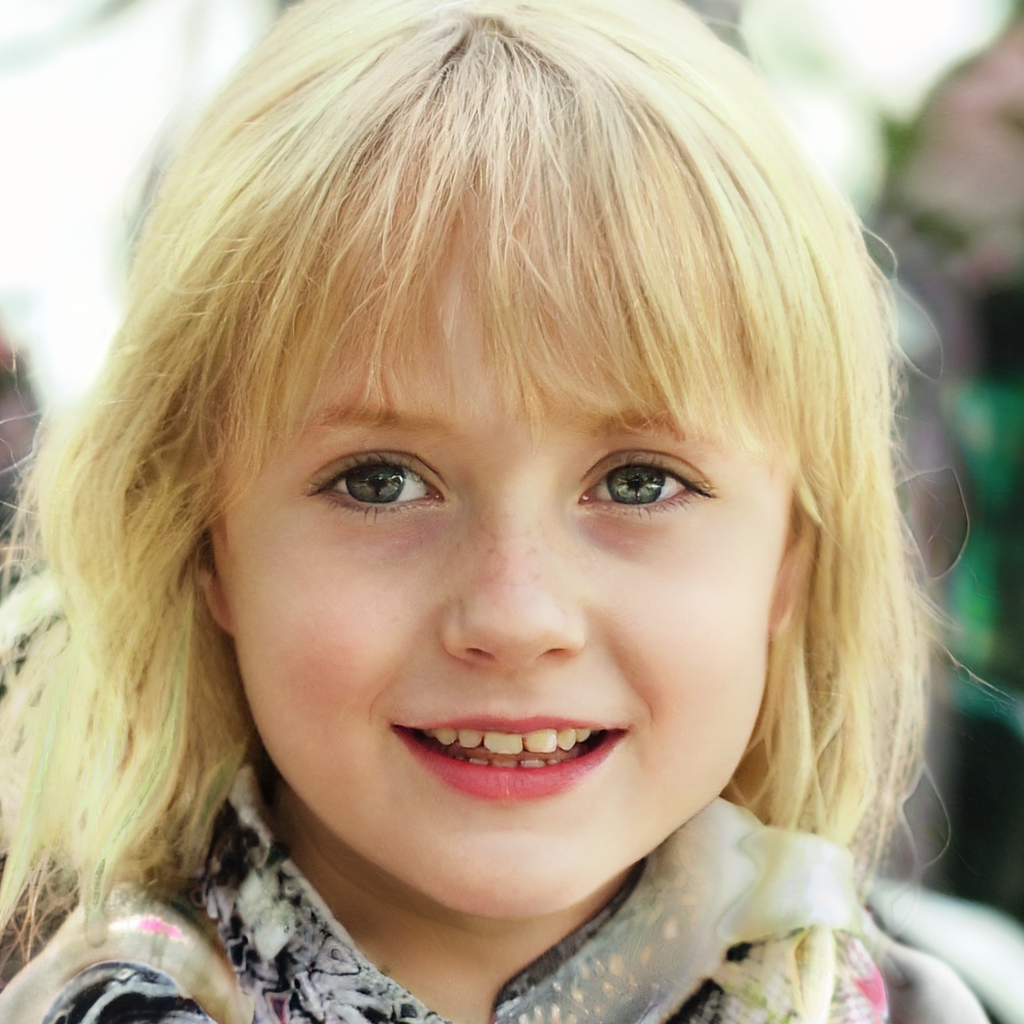}} 
~
\subfloat{\includegraphics[width=0.29\columnwidth]{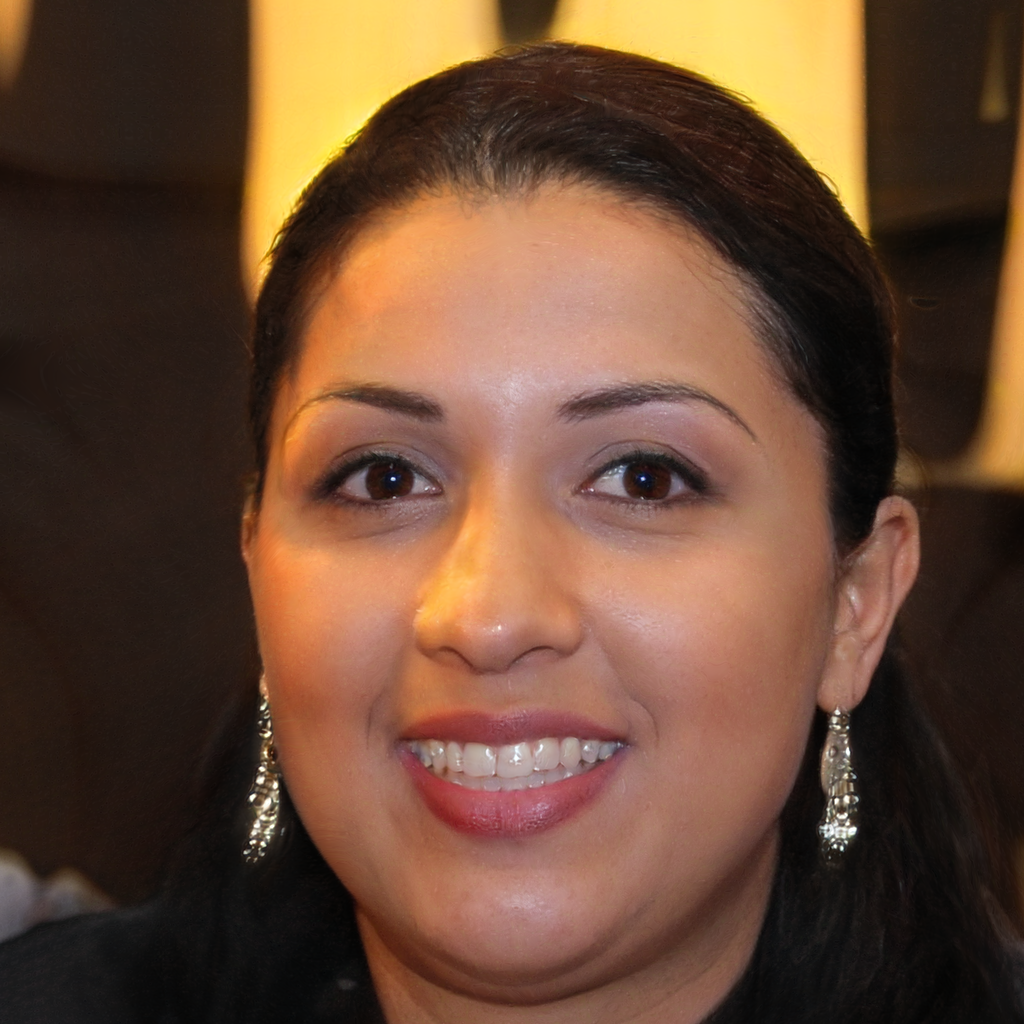}} \\

\subfloat{\includegraphics[width=0.29\columnwidth]{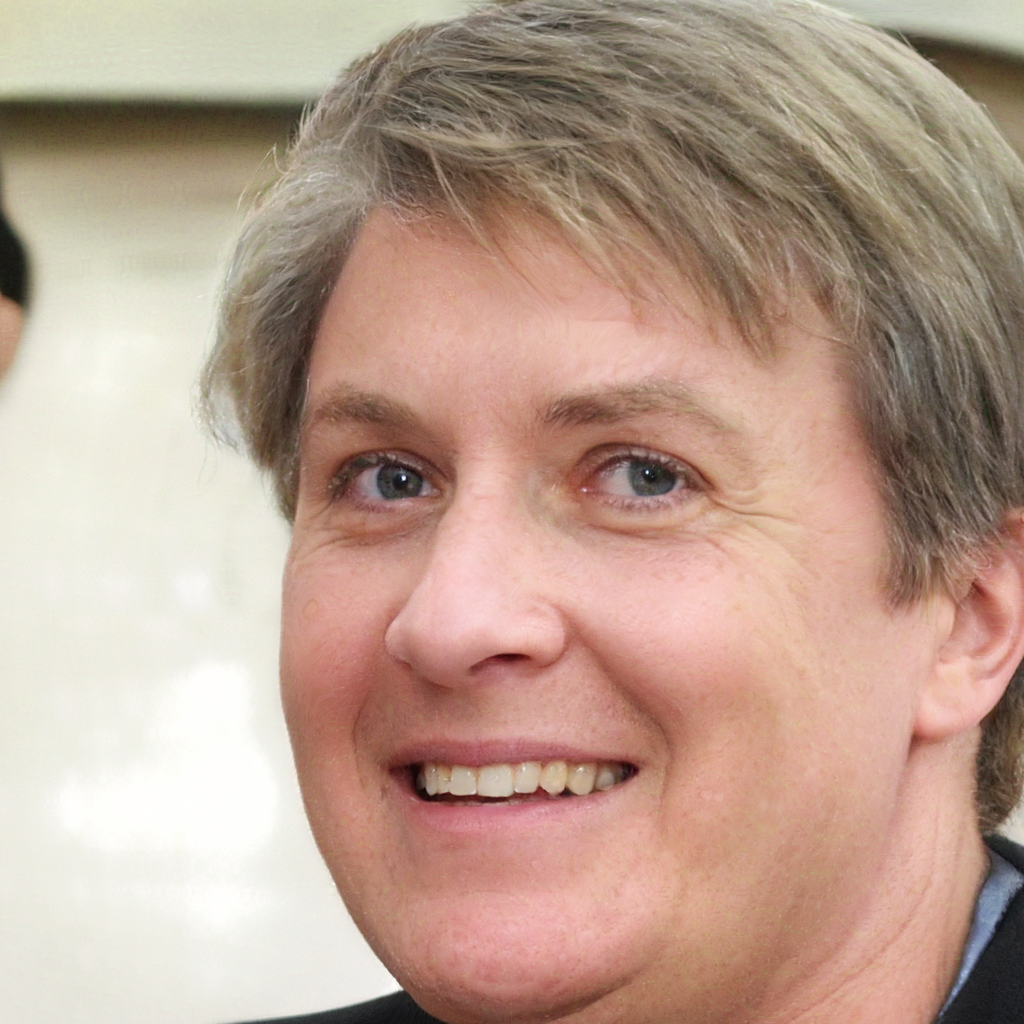}}
~
\subfloat{\includegraphics[width=0.29\columnwidth]{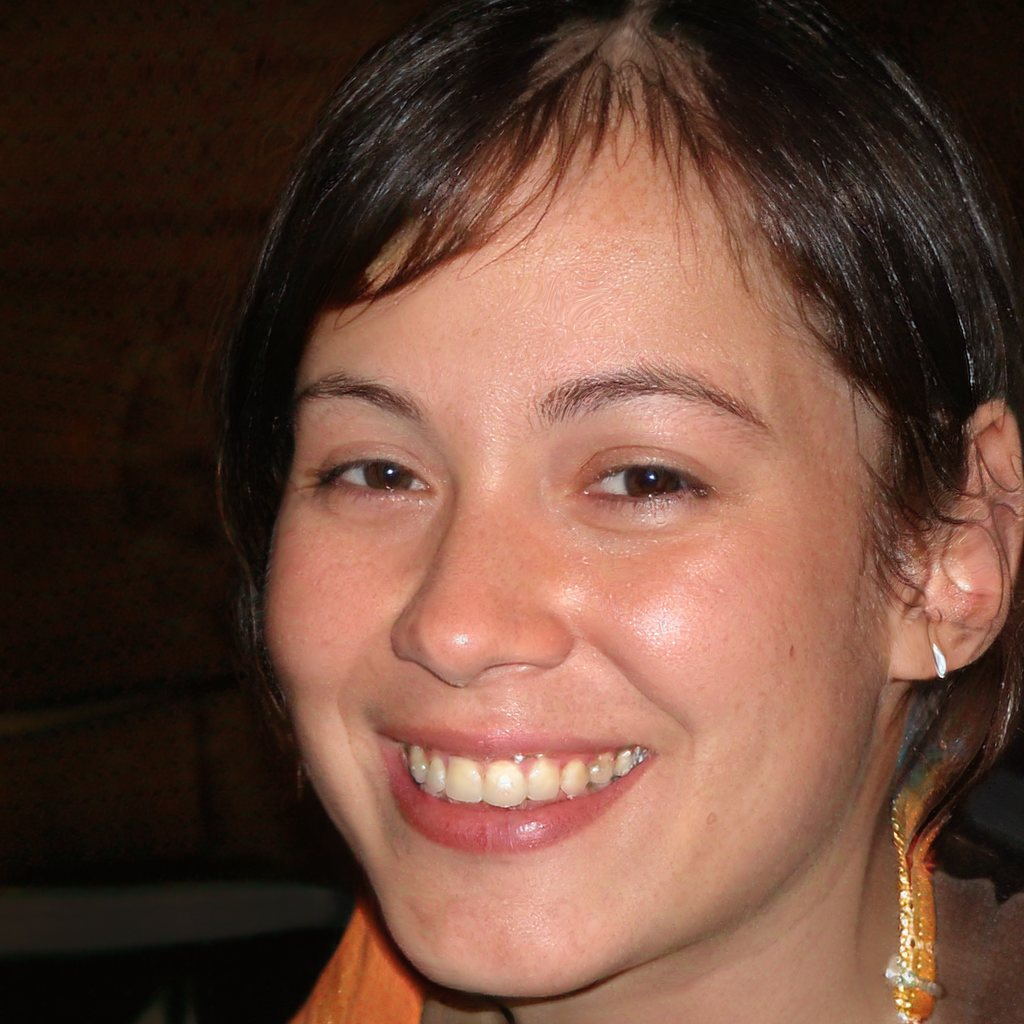}} 
~
\subfloat{\includegraphics[width=0.29\columnwidth]{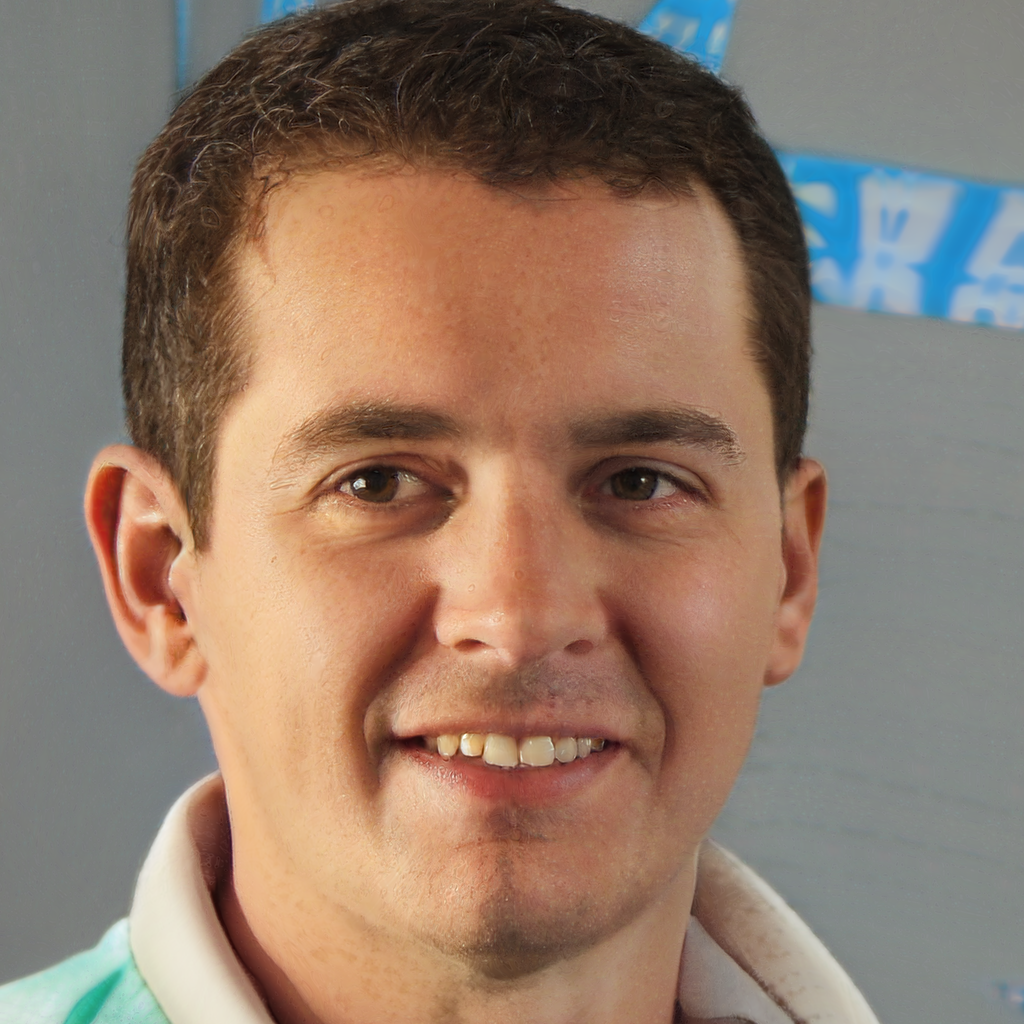}} \\

\caption{Examples of images generated using StyleGAN3 \cite{karras2021alias}. }
\label{fig:stylegan3}

\end{figure}

In fact, biometric data privacy is now considered a fundamental right and the collection and usage of such data is regulated under the law. For example, in 2017 the Indian supreme court ruled that privacy is a fundamental right of its residents and further emphasized that the protection of biometric data is of utmost importance. Recently, we have seen regulations such as the General Data Protection Regulation by the European Union that have brought the usage of biometric data under strict data protection laws. In the United States, similar steps have been taken. In California, the California Consumer Privacy Act (CCPA) and California Privacy Rights Act (CPRA) were put in place for this very purpose. In 2008, Illinois set forth the Biometric Information Privacy Act (BIPA), which has so far also been adopted by the states of Washington and Texas. Companies such as Facebook \cite{facebook}, Google \cite{google}, and Shutterfly \cite{shutterfly} have come under the scanner for their usage of facial images of their users under the BIPA law. On a national level, the National Biometric Information Privacy Act of 2020 (Senate Bill 4400) was introduced, if passed it would further strengthen the regulations on such data \cite{us-bill}. 

\begin{figure*}[htp]
\centering

\subfloat[Source]{\includegraphics[width=0.40\columnwidth]{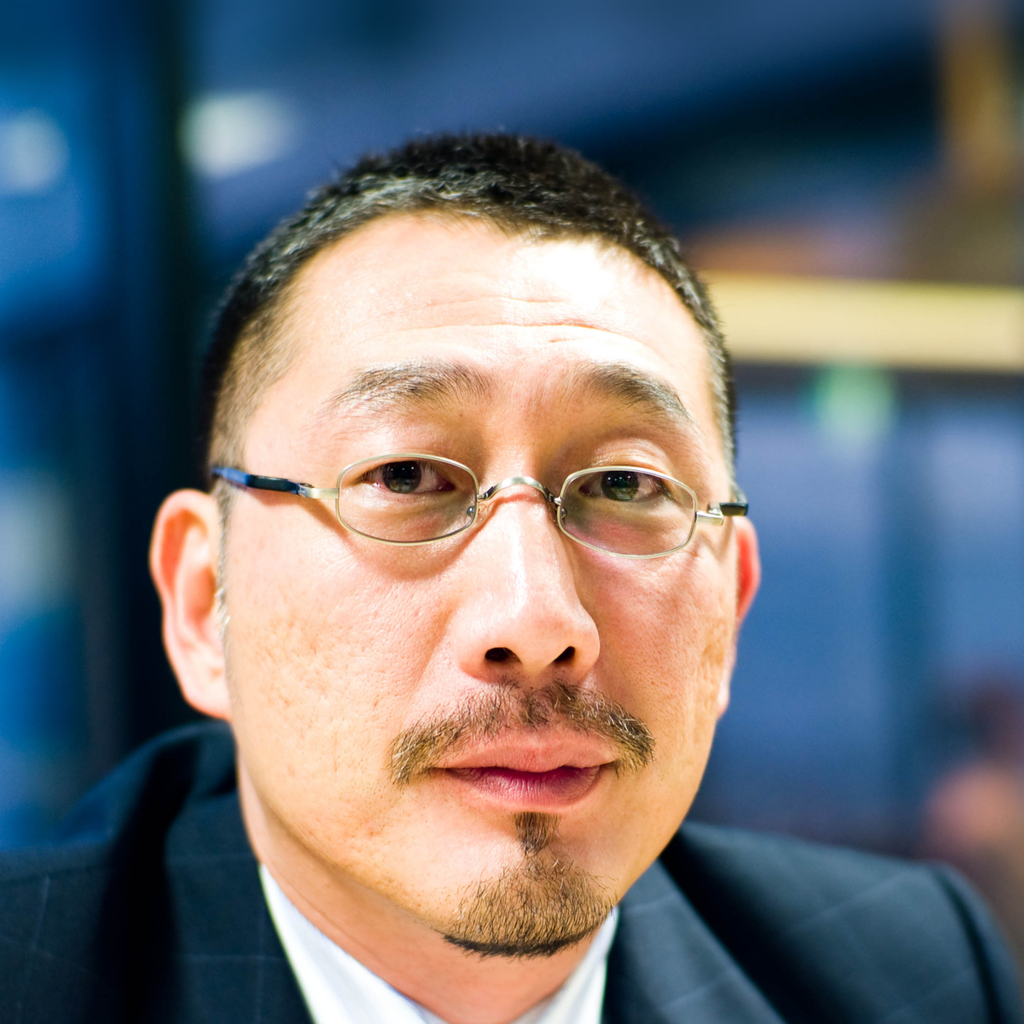}} 
~
\subfloat[Target]{\includegraphics[width=0.40\columnwidth]{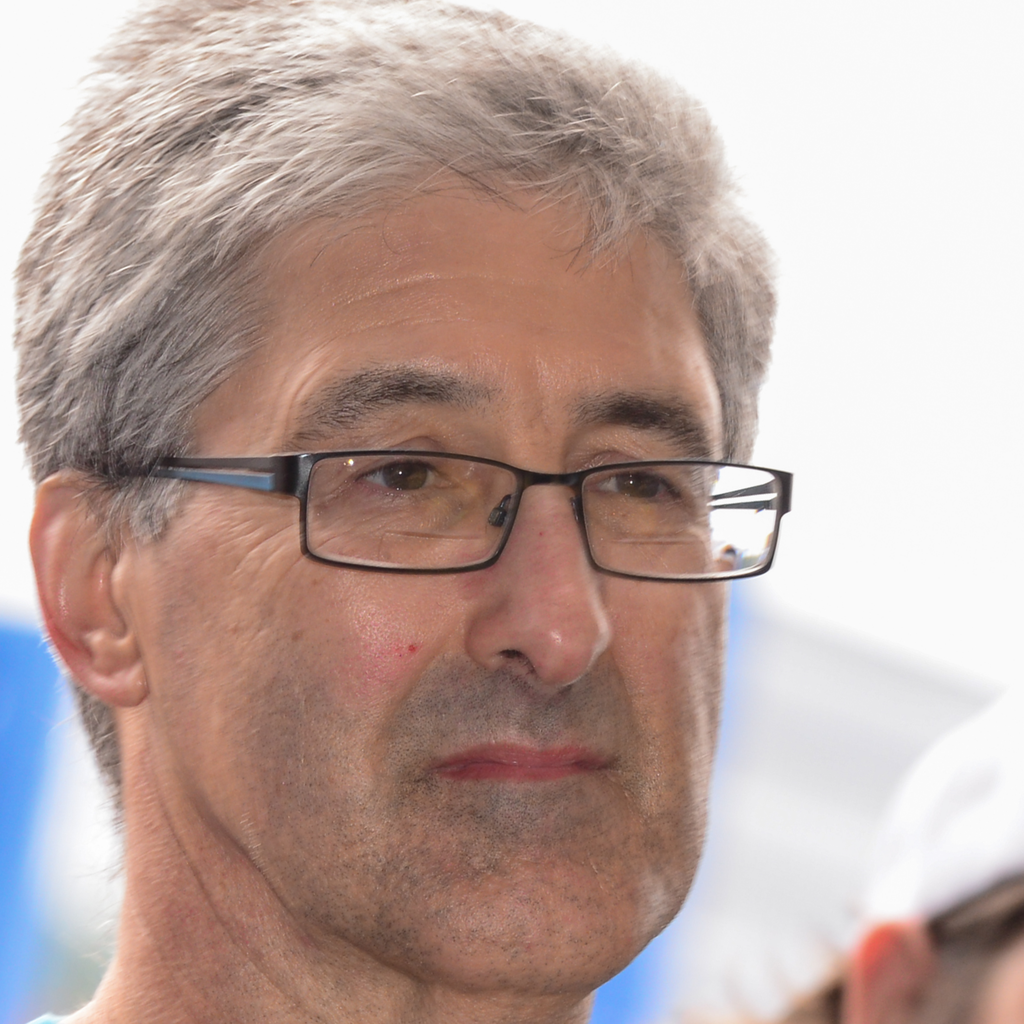}}
~
\subfloat[Swapped using SimSwap]{\includegraphics[width=0.40\columnwidth]{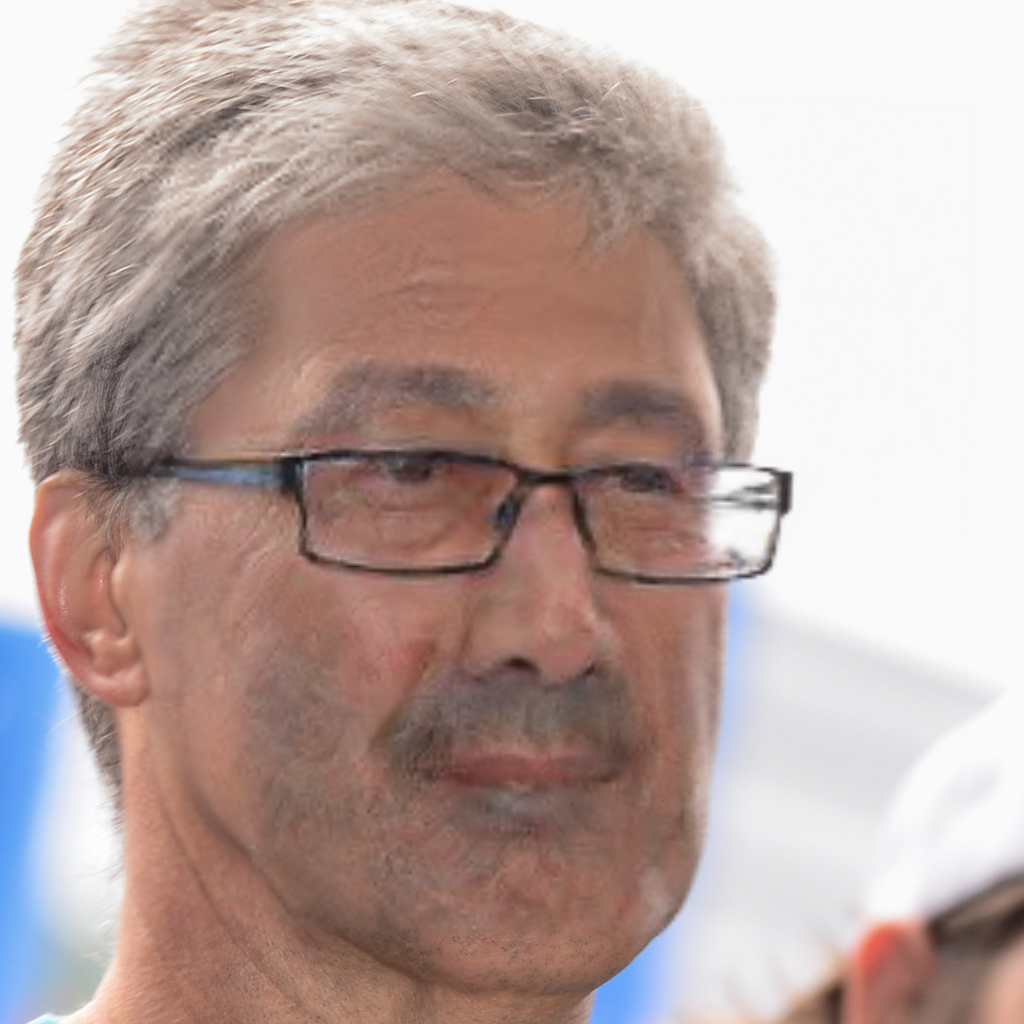}}
~
\subfloat[Swapped using SberSwap]{\includegraphics[width=0.40\columnwidth]{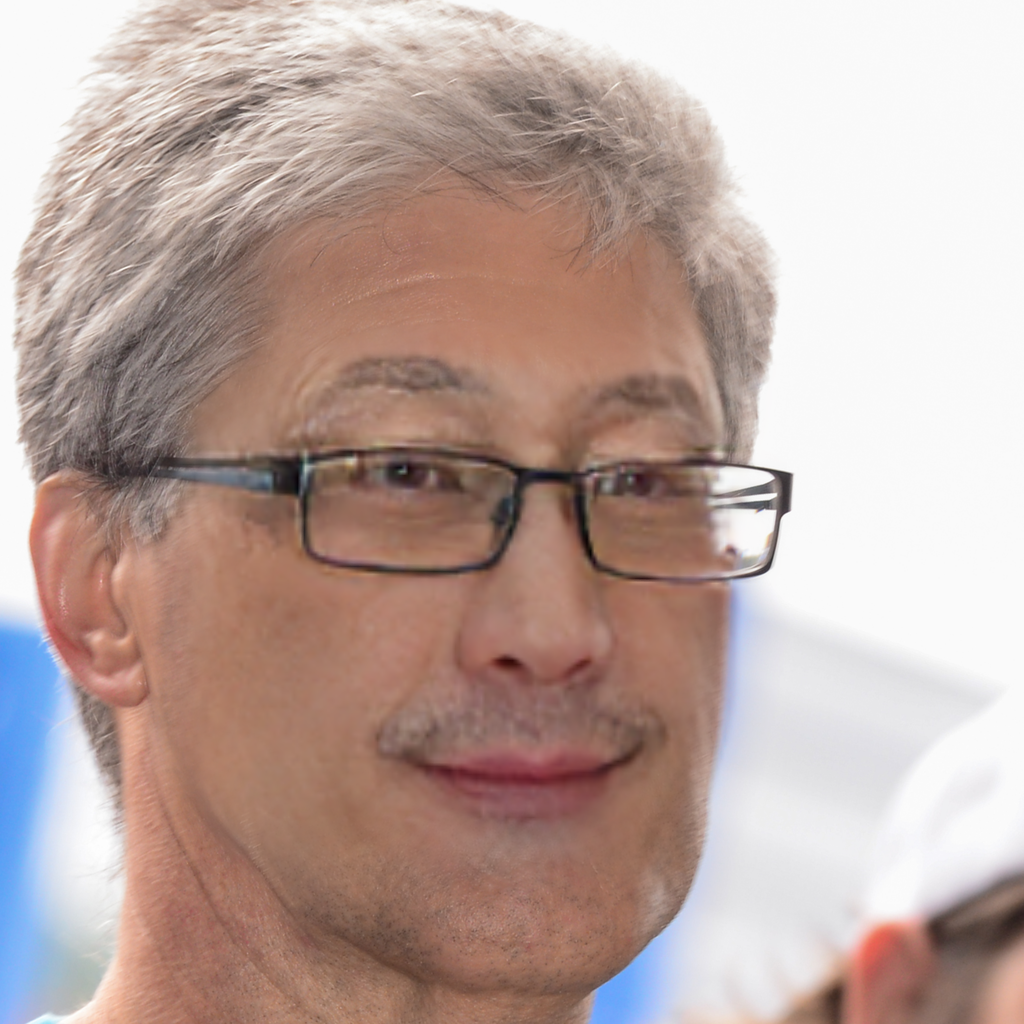}}
\\

\subfloat[Source]{\includegraphics[width=0.40\columnwidth]{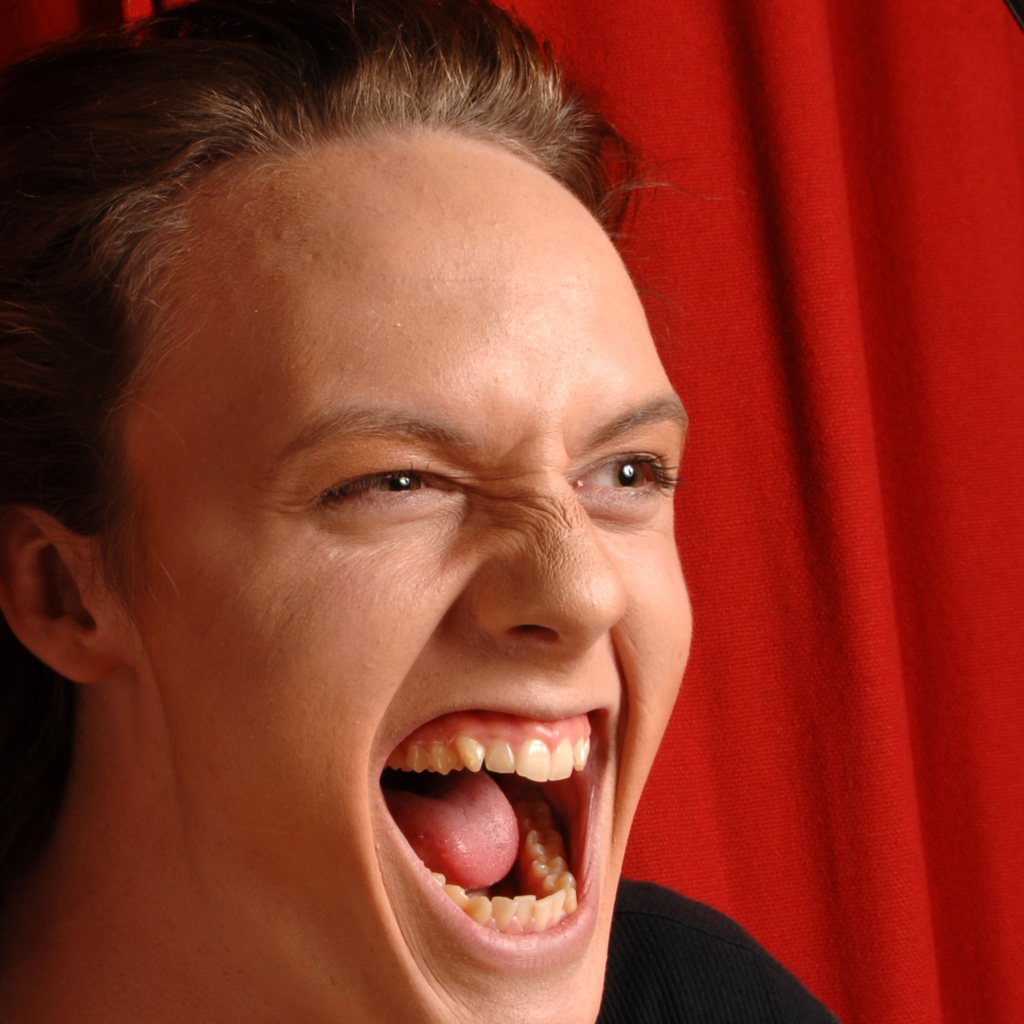}} 
~
\subfloat[Target]{\includegraphics[width=0.40\columnwidth]{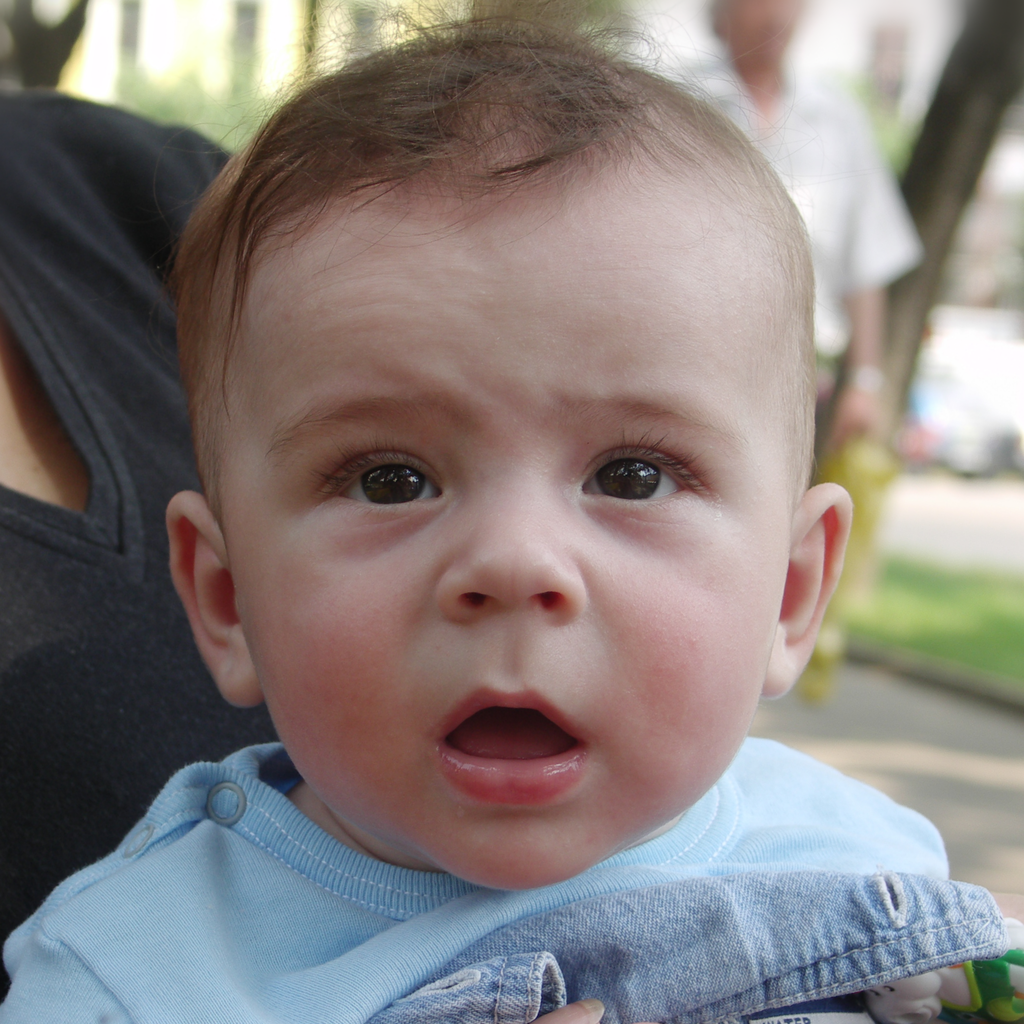}}
~
\subfloat[Swapped using SimSwap]{\includegraphics[width=0.40\columnwidth]{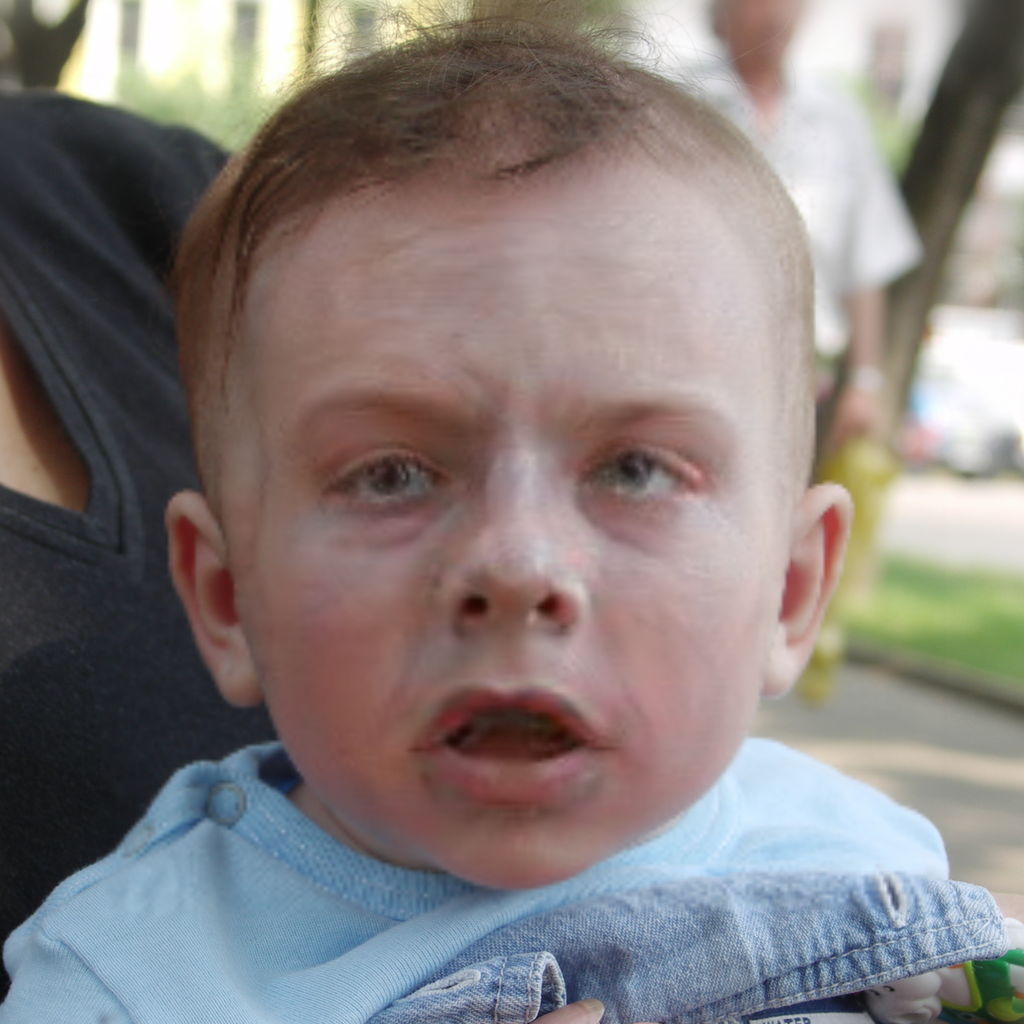}}
~
\subfloat[Swapped using SberSwap]{\includegraphics[width=0.40\columnwidth]{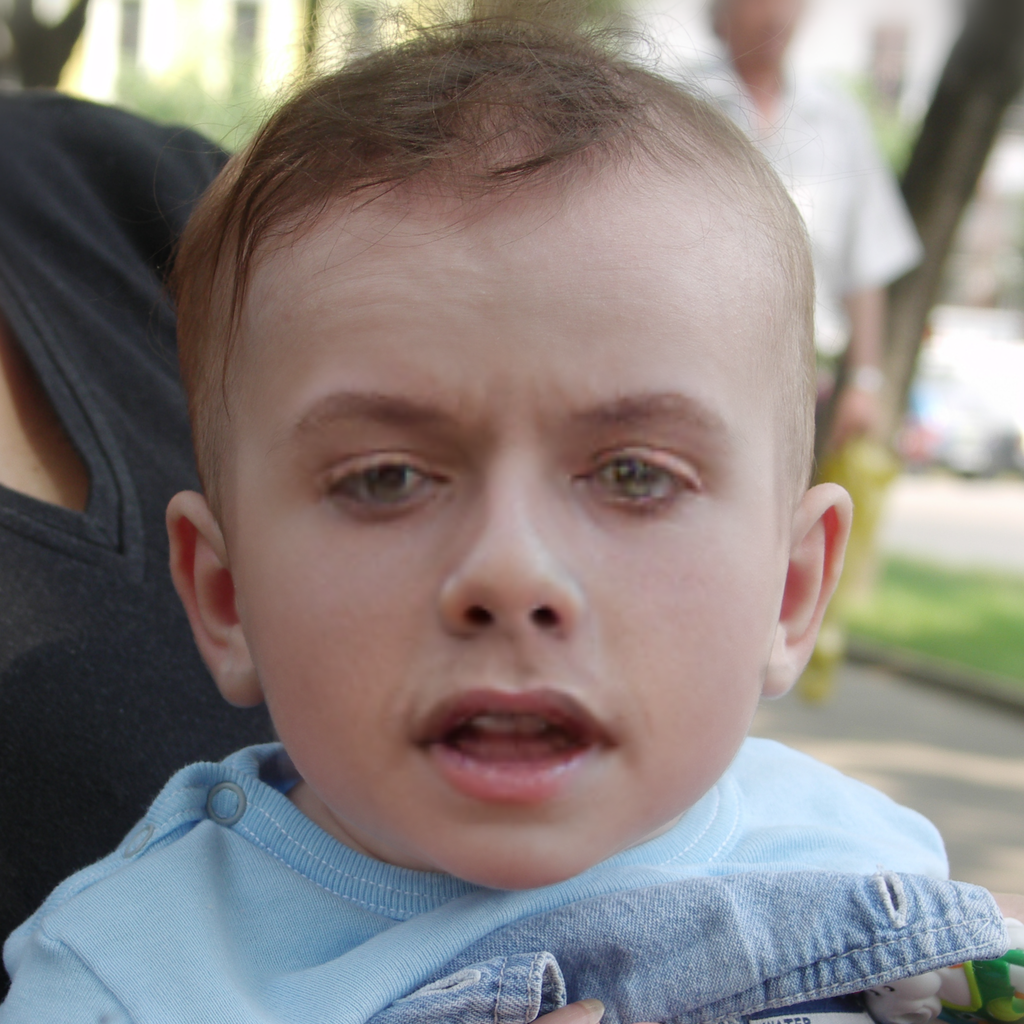}}\\

\caption{Examples of face swaps created using SimSwap and Sberswap. }
\label{fig:swapping}

\end{figure*}

There has also been an ongoing debate on whether researchers should be legally allowed to use publicly available biometric data and especially faces \cite{researcher-youtube}. While it is still not illegal to do so, it is often considered to be an ethical issue. This stems from the fact that the users appearing in these datasets are neither informed about their data being used nor has their consent been taken beforehand. If a user deletes their data from a public forum or wishes to do so, the machine learning models built using their data cannot be tracked and machine unlearning of this data which is no longer publicly available is close to impossible. Moreover, there is no way of tracking the datasets that are using data that the user no longer deems to be public. These ethical and legal conundrums have resulted in approaches for building vision models that move away from using data that contain biometric data of individuals.

In this paper, we propose a dataset-free approach for training a face-swap detection model that will help in bypassing such key privacy concerns by utilizing images of people that don't exist. We do this by generating fake identities and using these to create deepfakes or face swaps.

Since we create face swaps on the fly a major concern that arises with such an approach is that we assume that we have access to the face-swapping model in question. While this may seem to hamper its applicability in the real-world scenario, we show that this detection model built upon a face-swapper is generalizable to unseen face-swapping algorithms. We further emphasize this by showing that the detection model learns interpretable features through visualization using Captum \cite{kokhlikyan2020captum}. We observe that the model performs the best when it is fine-tuned with a small sample of real data to take care of the domain shift between synthetic and real data.  


Finally, we show that the deepfake detector trained on real-world data is biased against certain ethnicity. This has serious implications where deepfakes pertaining to this ethnicity is less likely to be detected. And as is often the case this arises due to biases existing in the dataset that is used to train the classification model. Our approach eliminates this dataset induced bias by using synthetic images. We show empirically in the paper that training a face swap detector using synthetic data helps in removing biases in the model architecture. A simple intuition is that synthetic datasets or GAN models give an additional level of control over the image we want. Even simply random sampling from the GAN latent space is powerful enough to reduce the bias significantly as we will show in this paper. 



\section{Related Work}


\subsection{Deepfake and Face Swapping Detection}

There has been a lot of recent interest in the detection of deepfakes and face-swapped images. Face-swapping methodologies can be divided into traditional face-swapping methods that use software such as Adobe Photoshop and the more recent GAN-based face-swapping models. In this paper, the focus is on the detection of the latter. Ding et al. \cite{ding2020swapped} proposed a transfer learning-based approach to detect face swaps that were created using Autoencoder GAN and Nirkin's method. The dataset that the authors created consisted of approximately 420,000 images. The frequency-domain has been used for the detection of face swaps in conjunction with a CNN-based network \cite{kohli2021detecting}. Guan et al. \cite{guan2021robust} proposed an approach to detect face swaps based on the inconsistencies in the 3D facial shape and appearance-based features. The authors tested the model on the FaceSwap and Deepfakes subset of the FaceForensics++ dataset. Other researchers have also proposed approaches that improve the generalizability of these classifiers such as Jiang et al. \cite{jiang2021practical}. However, their work requires the corresponding real face as a reference for prediction - which may not always be feasible to acquire. This also raises certain privacy concerns.

\subsection{Synthetic Datasets}

Generative adversarial networks have widely been used to create realistic synthetic imagery. In the past couple of years, GAN-generated images have become more and more realistic and have been successful in deceiving human beings. Now the question arises that since these images are indistinguishable from real images, can they also be used to train deep classifiers? Grosz et al. \cite{grosz2022spoofgan} showed that fingerprint spoofing classification benefits from the usage of synthetic GAN-generated fingerprints. They trained the spoofing classifier first only with generated images and further fine-tuned it using real images which shows better performance as compared to only training using real images. Jahanian et al. \cite{jahanian2021generative} used generated images as a source for multiview representation learning. They show that the learned representations are as good or sometimes even better than the ones learned using real data. Moreover, using contrastive learning, they were able to easily find positive pairs by generating images from nearby latent vectors.

Other benefits of using synthetic datasets have also been studied. Jaipuria et al. \cite{jaipuria2020deflating} showed that using synthetic data augmentation they were able to improve the model generalizability in the cross-dataset scenario by introducing more diversity in the training dataset. While these studies comprehensively show that there is a reasonable advantage to using synthetic data, they do not discuss the inherent biases introduced in the models due to class imbalance, and under-representation of particular ethnicities, genders, or racial groups. 

\section{Datasets}

In the paper we evaluate the trained face swap detection models on three datasets - Flickr Faces High Quality (FFHQ), CelebA-HQ, and the Amsterdam Dynamic Facial Expression Set (ADFES). 

The Flickr Faces HQ dataset \cite{karras2019style} contains 70,000 high quality 1024x1024 real closeup facial images. The images were scrapped from Flickr and thus have some distribution amongst various age groups, ethnicities, and backgrounds. The images were further processed by aligning and cropping them using dlib. We split the dataset into train and test sets with 30,000 images in the test set and the remaining 40,000 in the train set. For validation, we have utilized a small subset of images from the train test. 

The CelebA-HQ dataset \cite{DBLP:journals/corr/abs-1710-10196_celebahq} is higher quality version of the CelebA dataset of resolution 1024x1024. As the name suggests the dataset contains images of celebrities. The images have variable backgrounds and are diverse in terms of ethnic groups, age, and gender. We utilize the entire dataset for testing our models. Given the diversity in the dataset, we also use it for experimentation on understanding the biases present in the detection models.


The Amsterdam Dynamic Facial Expression Set (ADFES) \cite{fischer2012amsterdam} is a smaller dataset that contains 648 images from 22 models showcasing nine facial expressions: fear, anger, contempt, disgust, joy, sadness, pride, embarrassment, and surprise. The images in the dataset are of size 576x720 and thus had to be resized to 1024x1024 using bilinear interpolation. Moreover, the dataset contains only Northern European and Mediterranean models. The entire dataset is utilized for testing.

\section{Proposed Approach}

In this section, we describe the proposed pipeline of using synthetic images to train a face-swapping detection algorithm. To fairly evaluate the advantages of using synthetic images over using real images from the FFHQ dataset for training the classifier we have kept all other parameters consistent including preprocessing steps along with optimization parameters batch size and learning rate. 

\begin{table*}[]
\centering
\caption{Evaluation of the Xception trained only on synthetic data (row 1), real data from FFHQ dataset (row 2), and synthetic data with finetuning on FFHQ dataset (row 3) on the FFHQ, ADFES, and CelebA-HQ dataset. The classifier is trained to detect images from one face swapper and is tested in a cross-modal scenario on another. }
\begin{tabular}{llllll}
\toprule
Train Type        & Train Swap Model & Test Swap Model & FFHQ & ADFES & CelebA-HQ \\
\midrule
Trained on FFHQ   & Simswap     & Simswap    & 0.9973 & 1.0000 & 0.7437  \\
                  &             & Sberswap   & 0.9972 & 1.0000 & 0.7437 \\
                  \cline{2-6} 
                  
                  & Sberswap    & Simswap    & 0.9973 & 0.9979 & 0.9341 \\
                  &             & Sberswap   & 0.9974 & 0.9979 & 0.9341 \\

\midrule
Synthetic Data    & Simswap     & Simswap    &  0.9972 & 0.9979 &  0.8821 \\
                  &             & Sberswap   & 0.8802 & 0.9353  &  0.8509 \\
                  \cline{2-6}
                  
                  & SberSwap    & Simswap    & 0.9913 & 0.9916  & 0.8221 \\
                  &             & Sberswap   & 0.9921 & 0.9916  &  0.8236 \\
                  
\midrule
Finetuned on FFHQ & Simswap     & Simswap    &  0.9974 & 1.0000 & 0.9167 \\
                  &             & Sberswap   &  0.9971 & 1.0000 & 0.9164 \\
                  \cline{2-6} 
                  & Sberswap    & Simswap    & 0.9955 & 0.9979  & 0.9979 \\
                  &             & Sberswap   & 0.9973 & 1.0000 & 0.9877 \\
\bottomrule
\label{tab:results}
\end{tabular}
\end{table*}

\begin{table*}[]
\centering
\caption{Evaluation of the biases on the model trained with real data vs the models trained with synthetic data when testing on the CelebA-HQ dataset using the metrics Accuracy Difference (AD), Difference in Rejection Rate (DRR), and Difference in Acceptance Rate (DAR).  }
\resizebox{\linewidth}{!}{%
\begin{tabular}{llllllllllllll}
\toprule
Train Type        & Train Swap & Test Swap  & \multicolumn{3}{c}{Ethnicities} & \multicolumn{3}{c}{Gender}  & \multicolumn{3}{c}{Age}  \\
                &       Model            &       Model            & AD & DRR  & DAR & AD & DRR  & DAR & AD & DRR  & DAR \\ 
\midrule
Trained on    & Simswap     & Simswap    & 0.2233 & 0.2385 & 0.0181    & 0.1581 & 0.1555 & 0.0064 & 0.3384 & 0.5409 & 0.0069 \\
FFHQ                 &             & Sberswap   & 0.2573 & 0.2893  & 0.0181 & 0.1402 & 0.1236 & 0.0070 & 0.3388 & 0.6686 & 0.0103 \\
                  \cline{2-12} 
                  
                  & Sberswap    & Simswap    & 0.0933 & 0.1656 & 0.0160 & 0.0589 & 0.1030 & 0.0054 & 0.2828 & 0.9044 & 0.1429    \\
                  &             & Sberswap   & 0.1053 & 0.2172 & 0.0142 & 0.0542 & 0.0865 & 0.0056 & 0.2813 & 0.8983 & 0.1430 \\

                  
                  
\midrule
Finetuned  & Simswap     & Simswap    & 0.0729 & 0.1281 & 0.0164 & 0.0384 & 0.067 & 0.0054 & 0.3703 & 0.6646 & 0.0133\\
  on FFHQ                &             & Sberswap   & 0.0982 & 0.1923 & 0.0159 & 0.0331 & 0.0483 & 0.0047 & 0.3676 & 0.8624 & 0.0100 \\
                  \cline{2-12} 
                  & Sberswap    & Simswap    & 0.0180  & 0.0221 & 0.0184 & 0.0069 & 0.0151 & 0.0043 & 0.2223 & 0.9816 &  0.1250\\
                  &             & Sberswap   &  0.0198 & 0.0334 & 0.0154 & 0.0086 & 0.0117 & 0.0058 & 0.2223 & 0.9818 & 0.1250 \\
\bottomrule
\label{tab:bias}
\end{tabular}%
}
\end{table*}

\begin{figure}
    \centering
    \includegraphics[width=\columnwidth]{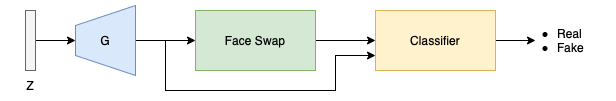}
    \caption{Overview of the proposed approach to detect face-swapped images using synthetically generated images.}
    \label{fig:diagram}
\end{figure}

The pipeline contains 3 major steps as shown in figure \ref{fig:diagram}. These are the generation of synthetic data, face swapping using the generated data, and finally classification of the face swaps.

\subsection{Data Generation}
We generated synthetic images using a pre-trained StyleGAN3\cite{karras2021alias} model. The StyleGAN3 model was trained on the Flickr-Faces High-Quality \cite{karras2019style} dataset to generate 1024x1024 closeup facial images. Figure \ref{fig:stylegan3} shows some examples of the generated images. Instead of creating a synthetic dataset \cite{jahanian2021generative} which would require a high amount of storage space we generate batches of images and use this for training. To ensure that the faces we generate are dissimilar we use a random vector generated using a different seed value for every batch. Moreover, we randomly select the truncation-psi value between $[0,1]$ using a normal distribution.

\subsection{Face Swapping}

The generated images from StyleGAN3 are further used for creating face swaps. The face swapper takes in a batch of GAN-generated images and generates $N-1$ number of swapped images by swapping the $i-th$ image with the $(i+1)-th$ image.  

We used two methodologies to swap faces - SimSwap \cite{chen2020simswap} and SberSwap \cite{chesakov2022new}. Both of these are GAN-based models that have been trained on the VGGFace2-HQ and VGGFace2 datasets respectively. 

The SimSwap architecture consists of three parts - the encoder, ID injection module (IIM), and the decoder. The ID injection module transfers the identity of the source image onto the target. The network is trained with an identity loss to ensure that the identity of the source image is generated. The network ensures that other facial attributes such as expression and gaze direction are transferred. 

The SberSwap algorithm is an improvement over the FaceShifter network \cite{chesakov2022new}. They make use of the AEI module combined with an adaptation of reconstruction loss proposed by SimSwap. They have an additional super-resolution block that makes the resulting images more realistic. They also added an eye loss, that was specifically made to ensure the consistency of the eye. The discrepancies in the eye and iris shape have been discussed in detail by previous researchers as being an easy way to spot deepfakes \cite{guo2022eyes}.

\subsection{Classification Network}

Finally, for the classification network, we use the Xception network \cite{chollet2017xception}. Authors of the FaceForensics++ dataset \cite{rossler2019faceforensics++} found this to be the best performing model on their dataset. We have used the same hyperparameters as in the original paper for consistency. 

The Xception network is trained for binary classification with a categorical cross-entropy loss function to distinguish between the face swapped and the original images. We changed the input and output classification layers from the original ImageNet model architecture. For training the classifier binary classification model, the synthetically generated images are considered as the "real class" and the face swapped images are considered as the "fake class". While the synthetic media isn't truly a "real class" and is often considered to be fake, in this particular case the deep learning model will learn distinguishing features between the two classes. Since both classes have synthetically generated images, the distinguishing features would come from the face swapper.

As a baseline model, we train the XceptionNet using only real images from the training set of the FFHQ dataset instead of using synthetically generated images. For fair comparison, only the training data has been changed, while keeping the model architecture and other training dynamics the same. The two models are trained using a batch size of 12 - which was the largest batch that could fit on an RTX-8000 GPU node. Due to the large size of the FFHQ dataset, we limited the number of steps per epoch to 2000 for both the models that were trained using real data and synthetic data. 



\section{Results}

\subsection{Detection Performance}

The classification performance of the models trained on real and synthetic data is tested both in the seen and unseen scenarios with respect to the face swap model and the dataset. We performed two sets of cross-modality experiments to test the efficiency of the proposed approach. First, we tested the model against an unseen face swap attack. For example, a model trained only on face swaps generated from SimSwap is tested on face swaps generated from SberSwap. Secondly, there might be some information leakage in the case of the FFHQ dataset through the StyleGAN3 model which was trained on this dataset, thus we test the model on two other datasets that have been collected in completely different settings and backgrounds. In fact images from the ADFES dataset had to be resized without maintaining the aspect ratio. A model that is able to generalize in such a scenario is immune to resizing-based image transformations. This is typically not the case with models that learn noise level differences and artifacts.

We observed that simply training on GAN-generated images led to good performance on the FFHQ and the ADFES dataset (~99\% accuracy for both) and even generalized to a different face-swapping method than it was originally trained on. However, the accuracy results for the CelebA-HQ dataset were lower as compared to the model trained on real FFHQ data. The model trained on real data with both the swapping and testing model being SberSwap had a detection accuracy of 93.41\% and for the same setting, the model trained on synthetic data had only 82.36\% accuracy. While there were also scenarios where the model trained on synthetic data performed better than the model trained on real data, this showed that there was still a need for overcoming the domain shift to perform better than the model solely trained on real data. For this purpose, we finetune the model trained solely on synthetic data with only 12,000 real images from FFHQ. This finetuned model performed better than the model solely trained on synthetic data in all combinations of datasets and face-swapping model - generalizing well across different datasets and face-swap technologies. 



\subsection{Bias}

We hypothesize that the models trained on synthetic data would be fairer or less biased. Since models trained on real data exhibit similar biases to the datasets they are trained on and in this case, the absence of a dataset allows an extra layer of control to mitigate bias. Our empirical results show that even simply random sampling images from a GAN model leads to significantly lesser bias in the resultant classification model. 

To test our hypothesis we utilize three metrics for evaluation of the post-training model biases - accuracy difference (AD), the difference in acceptance rate, and the difference in rejection rate. Accuracy difference is the maximum different between the classification accuracy of different facets in a set of demographics $ \{ d_1, d_2, ..., d_n \} \in D$ (equation \ref{eq:ad}). DAR and DRR have similarly been defined as the maximum difference in acceptance and rejection rates respectively (eq. \ref{eq:dar} and \ref{eq:drr}). Here acceptance rate is defined as the ratio of true positive predictions to the observed positives and the rejection rate is defined as the ratio of the true negative predictions to the observed negatives.

\begin{equation}
    AD = max_{i,j} | ACC_i - ACC_j| \forall i,j \in D
    \label{eq:ad}
\end{equation}

\begin{equation}
    DAR = max_{i,j} | AR_i - AR_j| \forall i,j \in D
    \label{eq:dar}
\end{equation}

\begin{equation}
    DRR = max_{i,j} | RR_i - RR_j| \forall i,j \in D
    \label{eq:drr}
\end{equation}

For this work, we have categorized the people based on their ethnicity/race, age, and gender. DeepFace model \cite{serengil2021lightface} was used for categorization of the image into these subgroups based on the predicted race, age, and gender. We have considered six ethnic groups of people - white, Asian, Indian, black, Latino Hispanic, and middle eastern based on the predictions of DeepFace. The people are classified into two genders - male and female and we have quantized the estimated age into buckets of 10 years. 


We observed significant improvement in the biases of the model fine-tuned on real data vs the model solely trained on real data. The results have been summarized in table \ref{tab:bias}. From this, we observe that the accuracy difference for the former is roughly 10 times lesser than the latter in most cases. Similarly, DRR is about half and DAR is almost the same. The only anomaly is in the case of biases in age groups where the model trained on real data is slightly better in terms of AD and DRR. However, it is important to note that particular age groups such as '10-20' and  '70-80' which had only 16 and 9 samples in the CelebA-HQ dataset have largely influenced these values. Thus, the major focus should remain on the values pertaining to the subcategories of ethnicity and gender where there are a sufficient number of samples per subcategory ($>500$).

\begin{figure}[t]
\centering
\subfloat{\includegraphics[width=0.5\columnwidth]{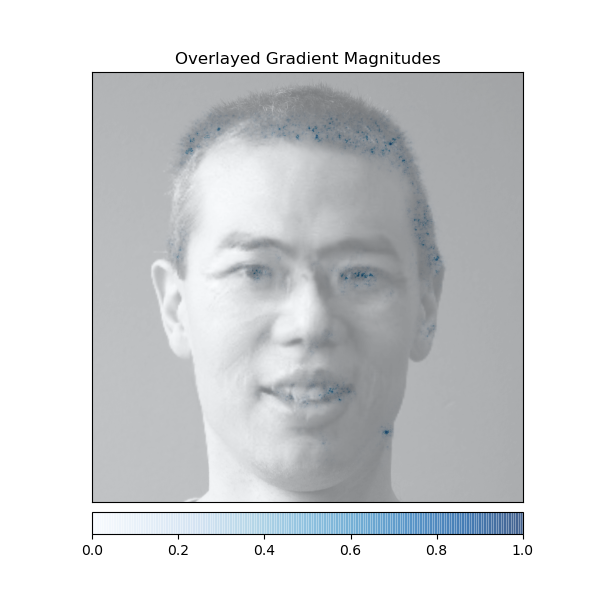}}
~
\subfloat{\includegraphics[width=0.5\columnwidth]{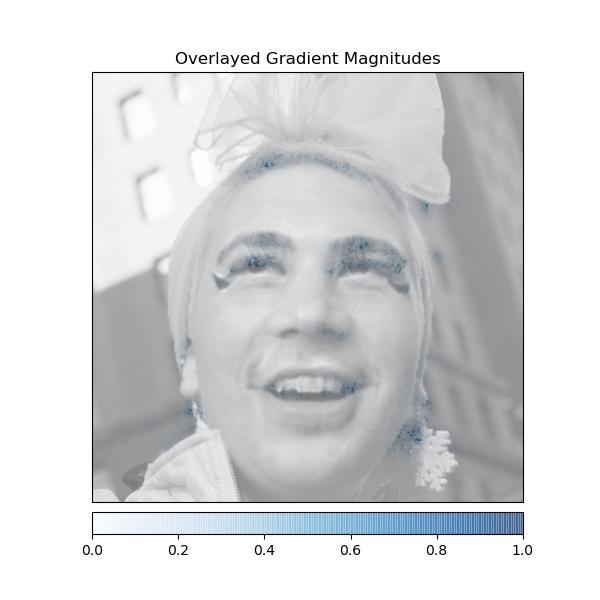}} 
\\

\subfloat{\includegraphics[width=0.5\columnwidth]{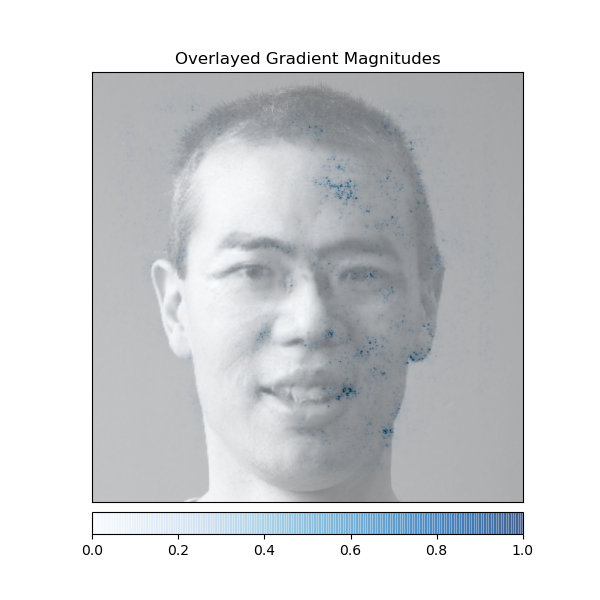}}
~
\subfloat{\includegraphics[width=0.5\columnwidth]{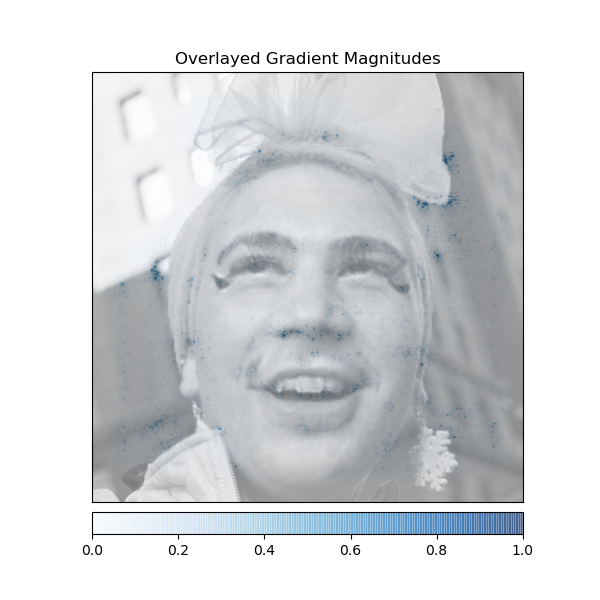}} 
\\

\caption{Observed Saliency plots for the XceptionNet models trained on synthetic data (top row) and real data (bottom row) showing the gradient magnitudes. }
\label{fig:occlusion}

\end{figure}

\subsection{Interpretability}
Finally, we analyze the model trained on synthetic data with the model trained on real data using Captum \cite{kokhlikyan2020captum} to observe the areas the model focuses on for making predictions. The objective behind this experiment is to observe whether these models are learning meaningful and human interpretable differences brought by the face swapper or if it is focusing on noise-level artifacts. We utilize two methodologies provided by Captum - Saliency and Occlusion. The former computes the gradients with respect to the class index and transposes the output for visualization. While for the latter a sliding window is used to iteratively occlude part of the image. The change in the model output is quantified and visualized. 

\begin{figure}[t]
\centering
\subfloat{\includegraphics[width=0.5\columnwidth]{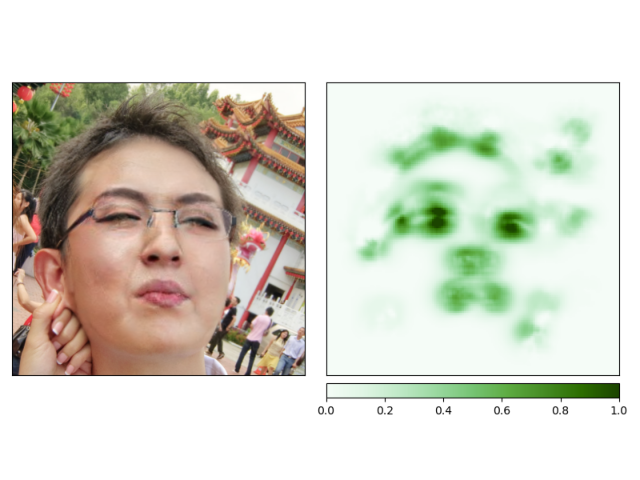}}
~
\subfloat{\includegraphics[width=0.5\columnwidth]{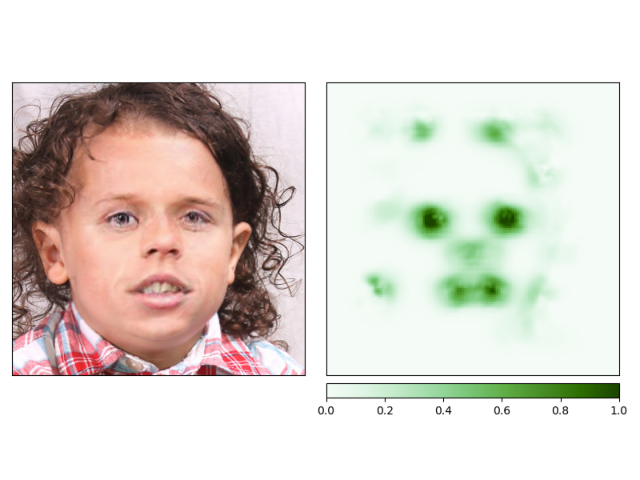}} 
\\

\subfloat{\includegraphics[width=0.5\columnwidth]{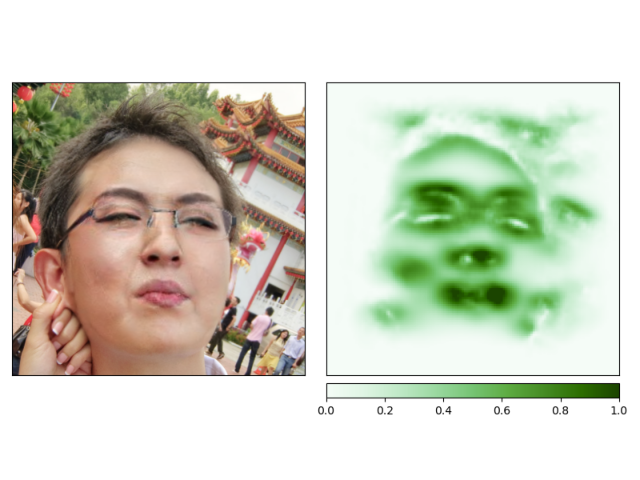}}
~
\subfloat{\includegraphics[width=0.5\columnwidth]{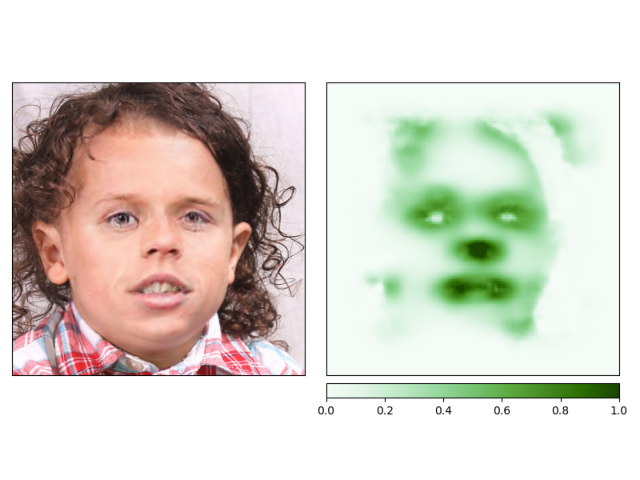}} 
\\

\caption{Examples regions that the model focuses on for making the decisions for both using synthetic data (top row) and real data for training (bottom row). }
\label{fig:saliency}

\end{figure}

Figure \ref{fig:saliency} depicts the Saliency plots for the two models. It can be seen that the model trained on synthetic data has more focused regions that clearly align with the human face attributes such as the eyes, nose, and mouth. However, for the model trained on real data, the higher valued gradients are more spread around the face and some background regions.

Similarly, for the Occlusion plots in figure \ref{fig:occlusion}, it can be observed that the model trained on synthetic data learns more interpretable and generalizable features. The model trained only on real data focuses on the entire face region along with some background regions while on the other hand, the model trained on synthetic data learns to specifically focus on facial attributes. 

Researchers in the past have shown that detectors tend to focus on noise level features generated by CNNs more easily when trained with real data \cite{jain2018detecting, wang2020cnn}. We hypothesize that the model trained on synthetic data learns more interpretable or higher-level image features as the synthetic images which are considered real for the classification model already contain some CNN-generated artifacts. These artifacts get overlayed on the ones introduced by the face-swap algorithm making it difficult to distinguish between them.

\section{Conclusion}

In this paper, we present an approach for detecting face swaps using synthetic data. We show that this dataless and privacy-aware approach of using faces that do not exist achieves competitive performance to models trained on real data. With some additional fine-tuning on real data, the model trained only on synthetic data can actually surpass the performance of the model trained on real data. In addition to the privacy and memory advantages of using synthetic datasets, we show that the final trained model is also less biased and learns more interpretable features. We measure the bias based on the performance of the model across the different races, genders, and age groups. Overall, we have shown several advantages to shifting from using real data for training such models to using synthetic data.

{\small
\bibliographystyle{ieee}
\bibliography{biblio}
}


\end{document}